\documentclass{amia}
\usepackage{graphicx}
\usepackage[labelfont=bf]{caption}
\usepackage[superscript,nomove]{cite}
\usepackage{color}

\usepackage{subfiles}

\newif\ifsubfile
\subfiletrue

\usepackage{soul}
\usepackage[color=yellow, textsize=small]{todonotes}
\usepackage{wrapfig}
\usepackage{multirow}
\usepackage{array}
\usepackage{booktabs}
\usepackage[hidelinks]{hyperref}
\usepackage{url}
\usepackage{enumitem}
\makeatletter
\renewcommand{\@biblabel}[1]{\hfill #1.}
\makeatother
\usepackage{amsmath}
\usepackage{bm}
\usepackage{dirtytalk}
\usepackage{subcaption}
\usepackage{makecell}

\usepackage{graphicx}
\usepackage{placeins}
\usepackage[export]{adjustbox}

\let\oldbibliography\thebibliography
\renewcommand{\thebibliography}[1]{%
  \oldbibliography{#1}%
  \setlength{\itemsep}{-0.1em}%
}

\begin{document}
\subfilefalse

% Learning objective: Learn about the extraction of radiological findings from radiology imaging reports and the automatic generation of structured semantic representations of these radiological findings, including the benefits, challenges, and methods.

\title{Extracting Radiological Findings With Normalized Anatomical Information Using a Span-Based BERT Relation Extraction Model}

\author{Kevin Lybarger, PhD$^1$, Aashka Damani,$^1$, Martin Gunn, M.B., Ch.B.$^1$, \\ \"{O}zlem Uzuner, PhD$^2$, Meliha Yetisgen, PhD$^1$}

\institutes{$^1$University of Washington, Seattle, WA, USA; $^2$George Mason University, Fairfax, VA, USA}

\maketitle

\noindent{\bf Abstract}

\textit{
Medical imaging is critical to the diagnosis and treatment of numerous medical problems, including many forms of cancer. Medical imaging reports distill the findings and observations of radiologists, creating an unstructured textual representation of unstructured medical images. Large-scale use of this text-encoded information requires converting the unstructured text to a structured, semantic representation. We explore the extraction and normalization of anatomical information in radiology reports that is associated with radiological findings. We investigate this extraction and normalization task using a span-based relation extraction model that jointly extracts entities and relations using BERT. This work examines the factors that influence extraction and normalization performance, including the body part/organ system, frequency of occurrence, span length, and span diversity. It discusses approaches for improving performance and creating high-quality semantic representations of radiological phenomena.
}

\section*{Introduction}

Radiology reports contain detailed descriptions of diverse clinical abnormalities based on radiologists’ interpretation of medical imaging. Although structured reports with semantic representations of medical concepts have been developed,\cite{rubin2017common} nearly all radiology reports convey findings through unstructured text.\cite{willemink2020preparing} Semantic representations of radiological findings could be automatically generated using natural language processing (NLP) information extraction techniques. These automatically derived semantic representations would enable a wide range of applications, including ground-truth labeling for artificial intelligence applications of medical images,\cite{zech2018natural} translation of reports into lay-language for patients, integration with clinical decision support,\cite{demner2009can} cross-specialty diagnosis correlation,\cite{filice2019deep} automated impression generation,\cite{wiggins2021natural} semantic searching of reports,\cite{gerstmair2012intelligent} and timely follow-up of recommendations.\cite{mabotuwana2019automated} We are currently conducting a large-scale clinical and economic analysis of incidental findings (incidentalomas) in radiology reports, focusing on six organ systems with the highest probability of incidental malignancy (thyroid, lung, adrenal glands, kidneys, liver, and pancreas). Incidentaloma identification requires the extraction of radiological findings and conversion of these findings to a structured semantic representation.

% Organ system specific citations
% thyroid,\todo{PMID: 22305454} lung,\todo{PMID: 22391408} adrenal glands,\todo{PMID: 31255202} kidneys, liver, and pancreas\todo{PMID: 18322885}

To develop data-driven extraction models, we designed an event-based annotation schema and annotated computed tomography (CT) reports. Each finding event is characterized by a trigger and set of attributes (assertion, anatomy, characteristics, size, size-trend, size count). In this paper, we use this gold standard corpus to explore the extraction of radiological findings with normalized anatomy information. We extract radiological findings and associated anatomies as a relation extraction task, where the extracted anatomies are normalized to a set of 56 pre-defined anatomy labels. We investigate this relation extraction task using Eberts and Ulges's Span-based Entity and Relation Transformer (SpERT).\cite{eberts2020span} SpERT is a state-of-the-art BERT model that jointly extracts entities and relations using span and relation output layers. As part of an ablation study, we use the gold anatomical spans to explore anatomy normalization, without extraction, to better understand the normalization task and the role of context. In this normalization experimentation, anatomy phrases are normalized at 0.89 F1 micro. In the extraction experimentation, finding spans are extracted at 0.83-0.92 F1, anatomy spans are extracted at 0.72-0.79 F1, and finding-anatomy relations are extracted at 0.63-0.72 F1. We explore the relationship between extraction performance, span length and diversity, and anatomy frequency. This work leverages state-of-the-art transformer-based extraction approaches and provides insight into the extraction of key finding and anatomy information from radiology reports.

\ifsubfile
\bibliography{mybib}
\fi

\section*{Related Work}

There is a large body of biomedical entity normalization work exploring the mapping of text spans to fixed vocabularies. A frequently explored ontology is the Unified Medical Language System (UMLS) \cite{bodenreider2004unified}, which includes the Systematized Nomenclature of Medicine-Clinical Terms (SNOMED CT) and RxNorm. The 2019 National NLP Clinical Challenges (n2c2)/Open Health NLP (OHNLP) shared task explored the normalization of pre-defined text spans in clinical notes to SNOMED CT and RxNorm concepts. Top performing teams used dictionary and string matching, cosine distance, retrieval and ranking, and deep learning, with the highest performing system utilizing deep learning.\cite{henry20202019} 

With large concept vocabularies, a frequently explored approach utilizes a two-step process, where a retrieval model identifies top candidate concepts and then a reranking model identifies the single best concept.\cite{Chen2020clinical, datta2020radlex, ji2020bert, Xu2020unified} Chen et al. normalized biomedical entities to SNOMED CT concepts, using knowledge sources to identify candidate concepts and an ensemble of machine learning approaches is used to identify target concepts.\cite{Chen2020clinical}  Datta et al. and Ji et al. explore biomedical entity normalization tasks using BM25 to identify top concept candidates and BERT to select the top concept.\cite{datta2020radlex, ji2020bert}  In the n2c2 challenge, Xu et al. uses a Lucene-based search that utilizes the UMLS and a BERT-based reranker.\cite{Xu2020unified} In this work, the anatomical concept vocabulary does not necessitate a retrieval model for identifying top candidates; however, we do utilize BERT-based models for identifying anatomy concepts. Tutubalina et al. investigate the normalization of medical concepts in social media posts to SNOMED CT concepts using a bidirectional recurrent neural network (RNN) and attention network to classify spans, incorporating semantic information from the UMLS.\cite{TUTUBALINA201893} Wang et al. explore a hierarchical anatomy normalization task with nine body parts (e.g. head and chest) and 41 sub-body parts (e.g. skull and brain).\cite{wang2019mapping} Wang et al. use Wikipedia as an anatomical knowledge source and explore different scoring functions for comparing anatomical entities to anatomical wiki pages. 

Recent work also explores both the extraction and normalization of biomedical entities, including anatomical spans. Tahmasebi et al. implement an unsupervised approach where anatomical phrases are identified using SNOMED CT and grammar-based patterns.\cite{tahmasebi2019automatic} Anatomical phrases are normalized by representing each phrase as the weighted sum of word embeddings and comparing the cosine similarity between anatomical phrases and target concept labels. This unsupervised approach outperforms a stacked bidirectional RNN and conditional random fields (CRF) model. Tahmasebi et al. identify 56 anatomical class labels corresponding to SNOMED CT IDs, which we use in this work. In a sequence tagging task, Zhu et al. predict eight anatomy classes (brain, breast, kidney, liver, lung, prostate, thyroid, and other) using a stacked bidirectional long short-term memory network (bi-LSTM) and CRF that incorporates sentence-level context vectors that are learned to predict the presence of each anatomical class in the sentence.\cite{zhu2019context} Zhu et al. experiments with incorporating sentence-level and report-level context and finds that incorporating report-level context improves classification performance. Similar to Zhu et al., we also explore the role of context in normalizing anatomical spans. Our work is differentiated from this prior work in that we extract anatomical information related to medical findings, and the anatomical phrases are normalized to a larger anatomy vocabulary. 

\ifsubfile
\bibliography{mybib}
\fi

\section*{Methods}

\subsection*{Data}

% 1.4
\begin{wrapfigure}{R}{3.9in}
 %\vspace{-0.7in}
  \begin{center}
    \includegraphics[width=3.9in]{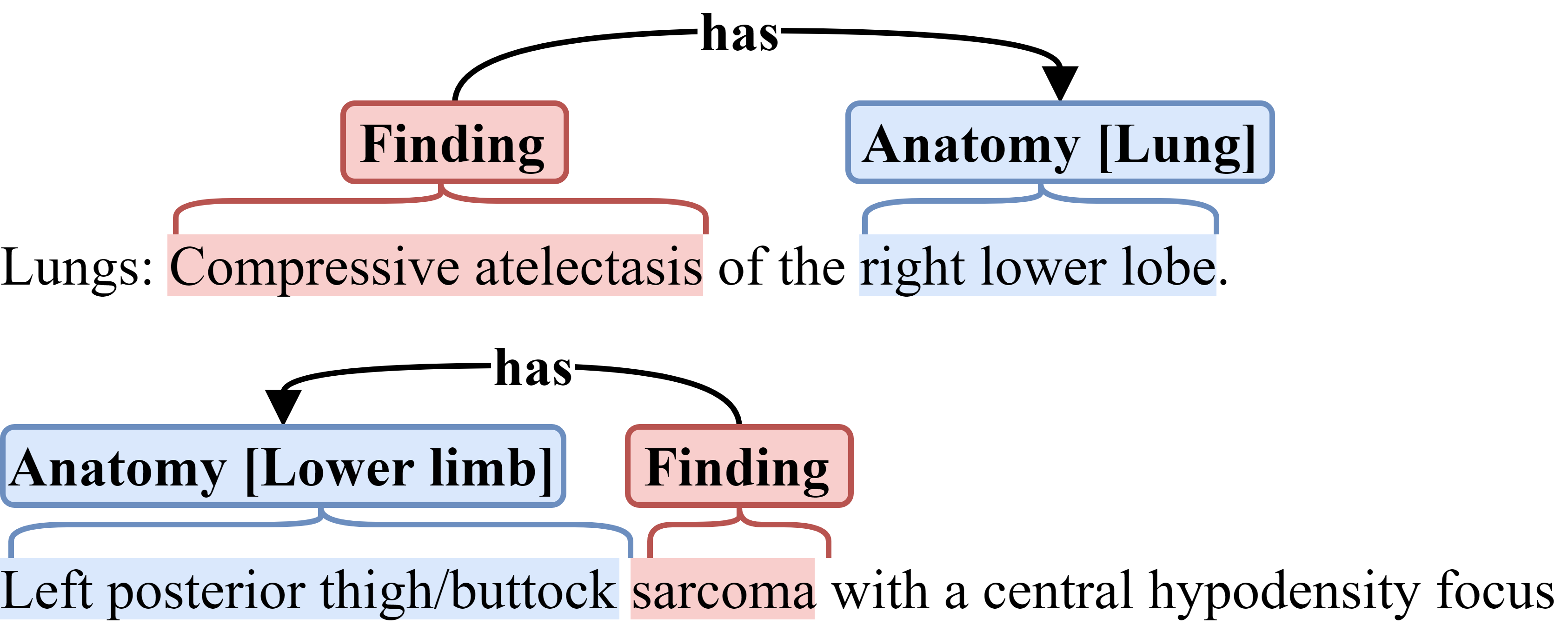}
  \end{center}
  \caption{Annotation example}
  \label{annotation_example}
\end{wrapfigure}

This work utilizes an annotated data set created by Lau, et al, \cite{lau2021new} which includes 500 randomly selected CT reports from an existing clinical data set from the University of Washington Medical Center and Harborview Medical Center. The data set includes 706,908 CT reports authored from 2008-2018. The annotated reports use an event-based annotation scheme to characterize two types of findings: (1) \textbf{lesion findings} (e.g. mass or tumor) and (2) \textbf{other medical problem findings} (e.g. fracture or lymphadenopathy). These findings are characterized across multiple dimensions, including assertion (e.g. present vs. absent), anatomy, count, size, and other attributes. The inter-rater agreement on event annotations in 30 notes is 0.83 F1. 

Although the corpus is annotated with several attributes related to findings, including lesions, this work focuses on the extraction of findings and the associated anatomical information. We collectively refer to the lesion findings and other medical problem findings as \textit{Finding}. Lau, et al.'s annotated corpus includes \textit{Anatomy} annotations without anatomy normalization labels.\cite{lau2021new} We augment the \textit{Anatomy} annotations to include the \textit{Anatomy Subtype} labels defined in Table \ref{anatomy_subtypes}. These \textit{Anatomy Subtype} labels are based on Tahmasebi et al.'s work identifying anatomical terms using unsupervised learning.\cite{tahmasebi2019automatic} The terms have associated SNOMED CT concept identifiers and represent all human organ systems, anatomic labels, and body regions. The \textit{Anatomy Subtype} labels normalize the \textit{Anatomy} spans, allowing the extracted finding and anatomy information to be more readily used in secondary use analyses.

\begin{table}[ht]
	\begin{minipage}{0.6\linewidth}
        \small
        \centering

%\begin{tabular}{lllllll}
%\toprule

%Abdomen          & Diaphragm	        & Heart	                 & Lymphatic$^{\star}$ & Ovary           & Prostate	      & Trach.       \\            
%Adrenal gland    & Digestive$^{\star}$ & Integumentary$^{\star}$ & Mediastinum	      & Pancreas	      & Retroperitoneal & Upper limb   \\     
%Back	           & Ear	              & Intestine	             & Mouth	            & Pelvis	        & Seminal vesicle & Urethra      \\  
%Bile Duct	       & Esophagus	        & Kidney	               & MSK$^{\star}$       & Penis	          & Spleen	        & Uterus       \\ 
%Bladder	         & Eye	              & Laryngeal	             & Nasal sinus	      & Pericardial sac	& Stomach	        & Vagina       \\ 
%Brain	           & Fallopian tube     & Liver	                 & Neck	              & Peritoneal sac	& Testis	        & Vas deferens \\       
%Breast	         & Gallbladder	      & Lower limb	           & Nervous$^{\star}$   & Pharynx	        & Thorax	        & Vulva        \\
%Cardio$^{\star}$  & Head	              & Lung	                 & Nose	              & Pleural sac	    & Thyroid	        & Whole body   \\     

%\bottomrule
%\end{tabular}

\begin{tabular}{llll}
\toprule
\multicolumn{4}{c}{\textbf{Anatomy Subtype labels}} \\ \midrule

Abdomen            & Gallbladder	          & Nasal sinus	     & Seminal vesicle \\
Adrenal gland      & Head	                  & Neck	           & Spleen	        \\
Back	             & Heart	                & Nervous$^{\star}$ & Stomach	      \\
Bile Duct	         & Integumentary$^{\star}$ & Nose	           & Testis	        \\
Bladder	           & Intestine	            & Ovary            & Thorax	        \\
Brain	             & Kidney	                & Pancreas	       & Thyroid	      \\
Breast	           & Laryngeal	            & Pelvis	         & Trach.         \\
Cardio$^{\star}$    & Liver	                & Penis	           & Upper limb     \\
Diaphragm	         & Lower limb	            & Pericardial sac	 & Urethra        \\
Digestive$^{\star}$ & Lung	                  & Peritoneal sac	 & Uterus         \\
Ear	               & Lymphatic$^{\star}$     & Pharynx	         & Vagina         \\
Esophagus	         & Mediastinum	          & Pleural sac	     & Vas deferens   \\
Eye	               & Mouth	                & Prostate	       & Vulva          \\
Fallopian tube     & MSK$^{\star}$           & Retroperitoneal  & Whole body     \\

\bottomrule
\end{tabular}

        \captionof{table}{\textit{Anatomy Subtype} labels. Abbreviated terms include Cardiovascular (Cardio), Musculoskeletal (MSK), and Tracheobronchial (Trach). $^{\star}$ indicates systems, like the \textit{Nervous System}.}
        \label{anatomy_subtypes}
	\end{minipage}\hfill
	\begin{minipage}{0.35\linewidth}
      %\centering
      \vspace{-0.3in}
      \includegraphics[width=2.0in]{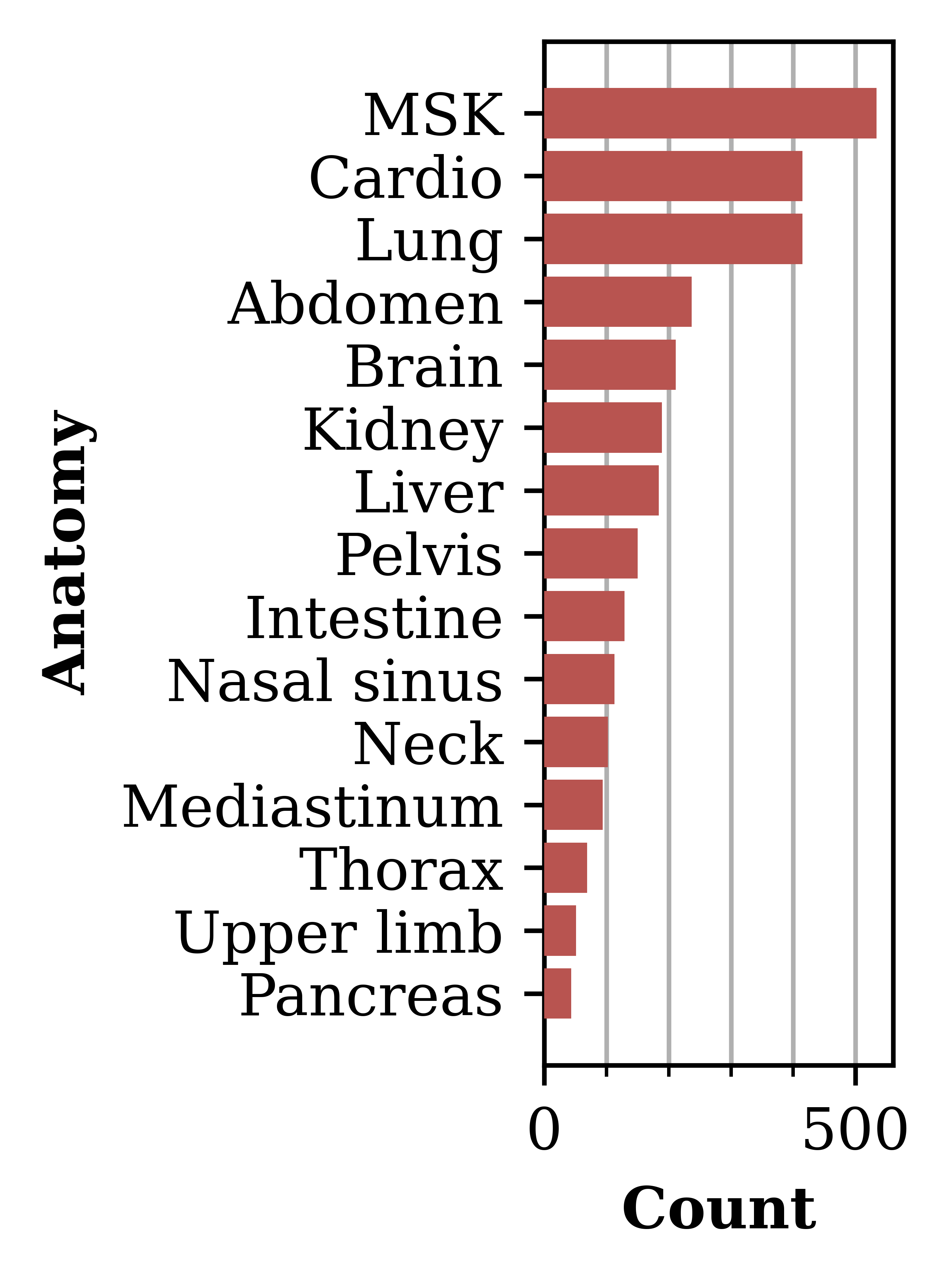}
      
      \vspace{-0.10in}
      \captionof{figure}{\textit{Anatomy Subtype} label distribution for training set}
      \label{anatomy_distribution}
	\end{minipage}
\end{table}

%\begin{table}[ht]
%    \small
%    \centering
%    \input{tables/anatomy_subtypes}
%    \caption{\textit{Anatomy Subtype} labels. Abbreviated terms include Cardiovascular (Cardio), Musculoskeletal (MSK), and Tracheobronchial (Trach). $^{\dag}$ indicates systems, like the \textit{Nervous System}.}
%    \label{anatomy_subtypes}
%\end{table}

%\begin{wrapfigure}{R}{2.2in}
%  %\vspace{-0.5in}
%  \begin{center}
%    \includegraphics[width=2.1in]{figures/anatomy_histogram_merged_subtype_train.png}
%  \end{center}
%  \caption{Anatomy distribution for training set}
%  \label{anatomy_distribution}
%\end{wrapfigure}

We approach this radiological information extraction task as a relation extraction task, where spans are identified, mapped to a fixed set of classes, and linked through relations. Figure \ref{annotation_example} presents example annotations. The entity types include \textit{Finding} and \textit{Anatomy}, although the phrases are not strictly noun phrases. Unlike a typical entity annotation, the \textit{Anatomy} entities include \textit{Anatomy Subtype} labels corresponding to the 56 anatomies defined in Table \ref{anatomy_subtypes}. We represent the \textit{Finding}-\textit{Anatomy} pairs as asymmetric relations, where the relation head is a \textit{Finding} entity and the tail is an \textit{Anatomy} entity. There is only a single relation type, \textit{has}, so the \textit{Finding-Anatomy} pairing can be interpreted as a binary classification task (connected vs. not connected).

The annotated corpus includes 500 CT reports, with 10,409 \textit{Finding} entities, 5,081 \textit{Anatomy} entities, and 6,295 \textit{Finding}-\textit{Anatomy} relations.\cite{lau2021new} There are more \textit{Finding}-\textit{Anatomy} relations than \textit{Anatomy} entities, because a given \textit{Anatomy} entity can be associated with multiple findings. The corpus includes approximately 19K sentences and 203K tokens and is randomly split into training (70\%), validation (10\%), and test (20\%) sets. Figure \ref{anatomy_distribution} presents the 20 most frequently annotated \textit{Anatomy Subtypes} in the training set. Musculoskeletal system (MSK), Cardiovascular system (Cardio), and Lung are the most frequent \textit{Anatomy Subtypes}, and there are many subtypes that occur infrequently or are absent from the data set. This skewed distribution is the result of randomly sampling the annotated corpus.

\subsection*{Information Extraction}

\begin{figure}[ht]
  \begin{center}
    \includegraphics[width=6.2in]{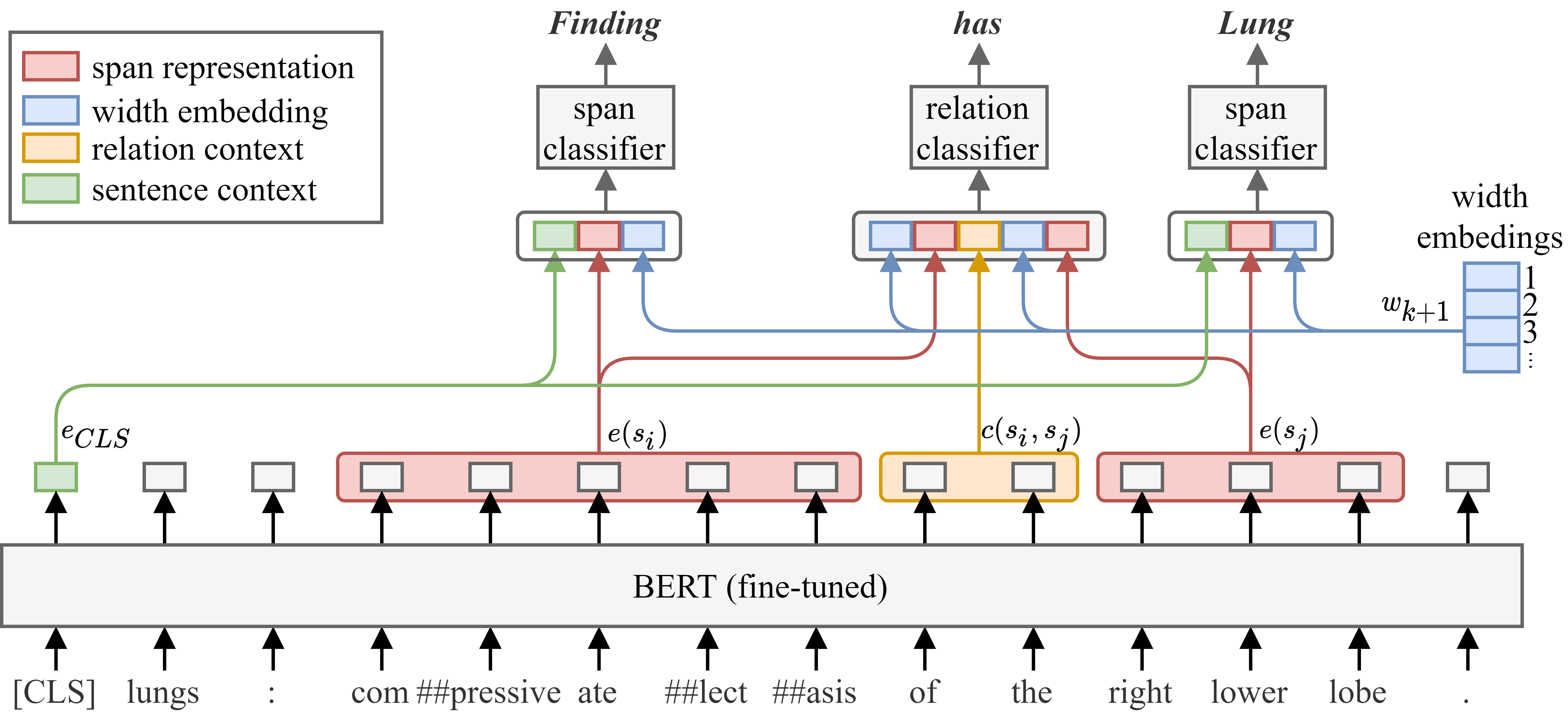}
  \end{center}
  \caption{SpERT framework}
  \label{spert}
\end{figure}

We extract the radiological findings and related anatomy using Eberts and Ulges's SpERT model.\cite{eberts2020span} SpERT jointly extracts entities and relations using a pre-trained BERT\cite{devlin2019bert} model with output layers that classify spans and predict the relations between spans. SpERT achieves state-of-the-art performance in three entity and relation extraction tasks, including open domain information extraction (CoNLL04), science information extraction (SciERC), and adverse drug event extraction (ADE).\cite{eberts2020span} The SpERT framework is presented in Figure \ref{spert}. 

\textbf{Input encoding:} Each sentence is tokenized and converted to BERT word pieces. BERT generates a contextualized representation for the sentence, yielding a sequence of word-piece embeddings $(\bm{e}_{CLS}, \bm{e}_1, \bm{e}_2, ... \bm{e}_t,..., \bm{e}_n)$, where $\bm{e}_{CLS}$ is the sentence-level representation associated with the $[CLS]$ token, $\bm{e}_t$ is the $t^{th}$ word piece embedding, and $n$ is the sequence length.

%\begin{figure}[ht]
%  \begin{center}
%    \includegraphics[width=5.6in]{figures/spert.png}
%  \end{center}
%  \caption{SpERT framework}
%  \label{spert}
%\end{figure}

\textbf{Span Classification:} The span classifier predicts labels for each span, $s = (t, {t+1}, ...{t+k})$, where the width of the span is $k+1$ word pieces. A learned matrix of span width embeddings, $\bm{w}$, is used to incorporate a span width prior in the classification of spans and relations. A fixed length representation of the $i^{th}$ span, $e(s_i)$, is created by max pooling the associated BERT embeddings and looking up the relevant span width embedding, as
\begin{equation}
\bm{e}(s_i) = MaxPool(\bm{e}_t, \bm{e}_{t+1}, ...\bm{e}_{t+k})\circ\bm{w}_{k+1},
\end{equation}
where $\circ$ denotes concatenation. The span classifier input for the $i^{th}$ span, $\bm{x}^s_i$, is the concatenation of the span embedding, $\bm{e}(s_i)$, and sentence-level context embedding, $\bm{e}_{CLS}$, as
\begin{equation}
\bm{x}^s_i = \bm{e}(s_i)\circ\bm{e}_{CLS}.
\label{span_classifier_input}
\end{equation}
The span classifier consists of a single linear layer, as 
\begin{equation}
\bm{y}^s_i = softmax(\bm{W}^s\cdot\bm{x}^s_i+\bm{b}^s).
\label{span_prediction_equation}
\end{equation}

For our task, the span classifier label set, $\Phi^s$, includes a $null$ label, \textit{Finding}, and the 56 \textit{Anatomy Subtypes} in Table \ref{anatomy_subtypes}: $\Phi^s=\{null, Finding, Abdomen, Adrenal \mbox{ } gland, \ldots, Whole \mbox { } body\}$ ($|\Phi^s|=58$). The $null$ label indicates \textit{no} span prediction. We experimented with several multi-layer, hierarchical span classifiers, where the first classification layer predicts the entity labels, $\{Finding, Anatomy\}$, and the second layer resolves the 56 \textit{Anatomy Subtype} labels for the \textit{Anatomy} spans. However, none of the hierarchical span classifier configurations outperformed the base SpERT model, so these hierarchical configurations are not presented. By directly predicting the \textit{Anatomy Subtypes}, the span classifier identifies and normalizes the \textit{Anatomy} spans. Only spans with a width less than a predefined maximum are included in modeling to limit time and space complexity.

\textbf{Relation Classification:} The relation classifier predicts the relationship between a candidate head span, $s_i$, and a candidate tail span, $s_j$, with input 
\begin{equation}
\bm{x}^r_{i,j} = \bm{e}(s_i)\circ\bm{c}(s_i, s_j)\circ\bm{e}(s_j),
\end{equation}
where $e(s_i)$ and $e(s_j)$ are the head and tail span embeddings and $\bm{c}(s_i, s_j)$ is the max pooling of the BERT embedding sequence between the head and tail spans. The relation classifier consists of a single linear layer, as
\begin{equation}
\bm{y}^r_{i,j} = softmax(\bm{W}^r\cdot\bm{x}^r_{i,j}+\bm{b}^r).
\end{equation}
For our task, the relation classifier label set, $\Phi^r$, includes a $null$ label and the relation types: $\Phi^r=\{null, has\}$  ($|\Phi^r|=2$). Only spans predicted to have a non-$null$ label are considered in the relation classification, to limit the time and space complexity of the pairwise span combinations. 

\textbf{Training:} The span and relation classifier parameters are learned while fine-tuning BERT. For each training batch, the cross entropy loss for each classifier is averaged, and the averaged loss values are summed using uniform weighting. The training spans include all the gold spans, $S^g$, as positive examples and a fixed number of spans with label $null$ as negative examples. The training relations include all the gold relations as positive in samples, and negative relation examples are created from all the entity pairs in $S^g$ that are not connected through a relation.

\textbf{Baseline:}
As a baseline for evaluating the performance of SpERT, we implement a multi-step BERT approach (BERT-multi) where entities are first extracted and then relations between entities are resolved. \textit{BERT-multi} is implemented by adding entity extraction and relation prediction layers to a single pretrained BERT model. For entity extraction, we implement a common BERT sequence tagging approach, where Begin-Inside-Outside (BIO) labels are predicted by a linear layer applied to the last BERT hidden state.\cite{Lee2019BioBERT} For evaluation, the word piece predictions are aggregated to token-level predictions by taking label of the first word piece of the token. For relation prediction, we implement a common BERT sentence classification approach, where relation predictions are generated by a linear layer applied to the $[CLS]$ encoding.\cite{Lee2019BioBERT} For each pair of predicted entities, a modified input sentence is created where the identified entities are replaced with special tags. For example, the first sentence in Figure \ref{annotation_example} would become, \say{Lungs: \textit{@Finding\$} of the \textit{@Lung\$}}. When enumerating candidate head-tail pairs, only \textit{Finding} entities are included as potential heads and only \textit{Anatomy Subtype} spans are included as potential tails. No such constraint is imposed in SpERT. Each training batch involves (i) generating sequence tag predictions and (ii) predicting relations for the identified spans, and the loss is backpropogated after both (i) and (ii).

\textbf{Experimentation:} The primary focus of this work is the extraction and normalization of anatomy information associated with findings. We use SpERT to extract \textit{Finding} and \textit{Anatomy} spans, normalize the \textit{Anatomy} spans to the \textit{Anatomy Subtypes}, and resolve \textit{Finding}-\textit{Anatomy} relations. The data set only includes anatomy annotations for anatomical information connected to findings, so not all anatomy phrases are annotated in the reports. 

We include normalization-only experimentation, where the \textit{Anatomy Subtype} labels are predicted for gold anatomy phrases. The normalization-only experimentation is incorporated to explore the difficulty of the anatomy normalization task separate from span extraction and investigate the role of context in anatomy normalization. This normalization-only experimentation uses the same input encoding and span classier as SpERT (see Equations \ref{span_classifier_input}-\ref{span_prediction_equation}). To investigate the role of context in anatomy normalization, we implement \textit{phrase-only} models where the input is the anatomy phrase (e.g. \say{\ul{right lower lobe}}) without any context and \textit{sentence context} models where each anatomy phrase is contextualized in the sentence in which it is located (e.g.\say{Lungs: Compressive atelectasis of the the \ul{right lower lobe}.}) Both normalization models use the gold labels to identify the anatomy phrases. 

%\begin{table}[ht]
%\begin{wraptable}{R}{3.7in}
    %\vspace{-0.4in}
%    \small
%    \centering
%    \input{tables/model_config}
%    \caption{Model configuration}
%    \label{model_config}
%\end{wraptable}
%\end{table}

%To investigate the role of the annotated \textit{Finding} entities in extracting the \textit{Anatomy} entities, we include \textit{Anatomy}-only entity extraction experimentation, where the gold \textit{Finding} entities and relations are not included in modeling.

Model architectures and hyperparameters were selected using the training and validation sets, and the final performance was evaluated on the withheld test set. Common parameters across all models include: pretrained transformer=\textit{Bio+Clinical BERT},\cite{alsentzer-etal-2019-publicly} optimizer=Adam, maximum gradient norm=1.0, and learning rate=5e-5. Normalization parameters include: dropout=0.05, batch size=50, and epochs=15. SpERT parameters include: dropout=0.2, batch size=20, epochs=20, learning rate warmup=0.1, weight decay=0.01, negative entity count=100, negative relation count=100, max span width=10, and maximum span pairs=1000. BERT-multi parameters include: batch size=50, epochs=20, dropout=0.2, negative relation count=100, and maximum span pairs=1000. To account for the variance associated with model random initialization, each model was trained on the training set 10 times and evaluated on the test set to generate a distribution of performance values. The mean and standard deviation (SD) of the performance values is presented (mean$\pm$SD). Significance is assessed using a two-sided t-test with unequal variance.

\textbf{Evaluation:} Performance is assessed using precision (P), recall (R), and F-score (F1). Each entity, $z$, can be represented as a double, $z=(s, \phi^s)$, where $s$ is the span $(t,..t+k)$ and $\phi^s$ is the span label in $\Phi^s$. Entity extraction performance is assessed using two sets of equivalence criteria: \textit{exact match} and \textit{any overlap}. Under the \textit{exact match} criteria, a gold entity, $z$, is equivalent to a predicted entity, $\hat{z}$, if the span and span label match exactly, as $(s \equiv \hat{s}) \land (\phi^s \equiv \hat{\phi}^s)$. Under the more relaxed \textit{any overlap} criteria, $z$ is equivalent to $\hat{z}$, if the there is at least one overlapping token in the gold and predicted spans and the span labels match, as $(s \mbox{ overlaps with } \hat{s}) \land (\phi^s \equiv \hat{\phi}^s)$. We include this \textit{any overlap} assessment, because the \textit{Anatomy Subtype} labels capture clinically relevant information, even if there are discrepancies in spans. In the example of Figure \ref{annotation_example}, the span \say{right lower lobe} is labeled as \textit{Anatomy} with \textit{Anatomy Subtype} Lung. If the span classifier predicts the span \say{lower lobe} to have the \textit{Anatomy Subtype} label Lung, the gold and predicted spans would not match, and the sidedness information associated with \say{right} would not be captured. However, a majority of the clinically relevant information would be captured, namely that the \textit{Finding} is associated with the Lung. Each relation, $r$, can be represented as a triple, $r=(z^h, \phi^r, z^t)$, where $z^h$ is the head, $\phi^r$ is the relation label in $\Phi^r$, and $z^t$ is the tail. A gold relation, $r$, and predicted relation, $\hat{r}$, are equivalent if $(z^h \equiv \hat{z}^h) \land (\phi^r \equiv \hat{\phi}^r) \land (z^t \equiv \hat{z}^t)$, where entity equivalence can be assessed using the \textit{exact match} or \textit{any overlap} criteria. 

\ifsubfile
\bibliography{mybib}
\fi

\section*{Results}

\subsection*{Normalization}

\label{normalization_results}

\begin{wraptable}{R}{3.2in}
        \vspace{-0.2in}
        \small
        \centering

\begin{tabular}{ll}
\toprule
\textbf{Model}   & \textbf{F1 micro} \\ \midrule
phrase only      & 0.86$\pm0.003$          \\
sentence context & 0.89$\pm0.004^\dag$          \\ \bottomrule
\end{tabular}

        \caption{Anatomy normalization performance on test set (mean$\pm$SD) for 10 models (1,153 phrases). $^\dag$indicates best performance with significance ($p < 0.05$).}
        \label{normalization_performance}
\end{wraptable}

This section presents the normalization results where \textit{Anatomy Subtype} labels are predicted for gold anatomy phrases. Table \ref{normalization_performance} presents the anatomy normalization performance on the withheld test set averaged across the 10 randomly instantiated models for each input configuration: \textit{phrase only} and \textit{sentence context}. The F1 scores in Table \ref{normalization_performance} are micro averaged across the 56 \textit{Anatomy Subtype} labels. The \textit{phrase only} model achieves relatively high performance, indicating a high proportion of the anatomical phrases include strong cues for normalization. The inclusion of the sentence context improves normalization performance from 0.86 F1 to 0.89 F1 with significance ($p<0.05$), indicating there are some ambiguous anatomy phrases that require intra-sentence context to resolve. For example, the term \say{cervical} can be related to the neck or the uterus, and sentence context is needed to resolve ambiguity. Early experimentation with context beyond the sentence of the anatomy phrase did not improve performance.

\begin{wraptable}{R}{3.2in}
        \vspace{-0.2in} 
        \small
        \centering

\newcolumntype{M}[1]{>{\centering\arraybackslash}m{#1}}
\newcolumntype{R}[1]{>{\centering\arraybackslash}m{#1}}

\begin{tabular}{m{0.5in} m{0.5in} R{0.6in}}
\toprule
\textbf{Gold}       & \textbf{Predicted}        & \textbf{Avg. freq.} \\ \midrule
Cardio       	    & MSK   	                & 18.6    \\
Abdomen	            & Intestine	                &    5.0    \\
Cardio      	    & Pelvis	                & 5.0    \\
MSK     	        & Thorax	                & 4.5   \\
Eye	                & MSK    	                & 4.3    \\
Intestine	        & Abdomen	                &   3.5 \\
MSK     	        & Cardio    	            & 3.4   \\
Head	            & Neck	                    & 2.8   \\
Lung	            & Thorax	                & 2.7   \\
Thorax	            & Abdomen	                &  2.5   \\
%Mediastinum	        & Thyroid	                &   2.3 \\
%Bile Duct	        & Gallbladder               & 	2.1   \\
%Integumentary sys.  & Lower limb                & 	2     \\
%Thorax	            & MSK sys.	                &    2   \\
%Esophagus	        & Thorax	                &  2     \\
\bottomrule
\end{tabular}

        \caption{Most confused \textit{Anatomy Subtypes} for \textit{sentence context} models on the test set, averaged across 10 models.}
        \label{norm_confusion}
\end{wraptable}

%\begin{table}[!htb]
%    \begin{minipage}{0.48\linewidth}
%        \small
%        \centering
%        \input{tables/normalization_perf.tex}
%        \caption{Anatomy normalization performance on withheld test set, as mean and SD for 10 trained models (1,150 anatomy phrases). $^\dag$indicates best performance with significance ($p < 0.05$).}
%        \label{normalization_performance}
%    \end{minipage}%
%    \hfill
%    \begin{minipage}{0.48\linewidth}
%        \small
%        \centering
%        \input{tables/norm_confusion}
%        \caption{Most confused \textit{Anatomy Subtype} labels for \textit{sentence context} models. Confusion frequency averaged across 10 trained models.}
%        \label{norm_confusion}
%    \end{minipage} 
%\end{table}

Table \ref{norm_confusion} presents the most frequently confused \textit{Anatomy Subtypes}, averaged across the \textit{sentence context} model predictions. We omit the full confusion matrix because of the high number of labels and sparsity of the matrix. In general, organs and body regions are the most confusable anatomy subtypes as either could be applied. Cardio and MSK are among the most frequently confused labels, with 53\% of all errors involving Cardio or MSK labels as the gold or predicted labels. Cardio and MSK labels are organ systems that extend throughout the body and therefore overlap with body region labels. Moreover, these labels are the most frequent in the data set. Other frequently confused labels include co-located body parts and organ systems, like Abdomen-Intestine and Head-Neck.

\subsection*{Entity and Relation Extraction}

\label{extraction_results}

%\begin{wraptable}{R}{4.5in}
\begin{table}[ht]
    \setlength\tabcolsep{3.8pt}
    \small
    \centering

    \begin{subtable}[h]{6.5in}    
        \small
        \centering

\begin{tabular}{m{1.35in} cccccccccccc}
\toprule
\multirow{3}{*}{\textbf{Span label}}  & \multirow{3}{*}{\textbf{\# gold}} & \multicolumn{7}{c}{\textbf{SpERT}}                                                             &  & \multicolumn{3}{c}{\textbf{BERT-multi}}    \\ \cmidrule[\heavyrulewidth]{3-9} \cmidrule[\heavyrulewidth]{11-13}
                                      &                                   & \multicolumn{3}{c}{\textbf{exact}}            & & \multicolumn{3}{c}{\textbf{overlap}}         &  & \textbf{exact}       & & \textbf{overlap}   \\ \cmidrule{3-5} \cmidrule{7-9} \cmidrule{11-11} \cmidrule{13-13}
                                      &                                   & \textbf{P} & \textbf{R} & \textbf{F1}    & & \textbf{P}  & \textbf{R} & \textbf{F1}  &  & \textbf{F1}     & & \textbf{F1}      \\ \midrule
Finding                               & 2,122                             & 0.82       & 0.84       & 0.83$\pm0.004^\dag$        & & 0.91        & 0.92       & 0.92$\pm0.004^\dag$      &  &  0.79$\pm0.005$        & & 0.91$\pm0.003$                \\
Anatomy                               & 1,153                             & 0.75       & 0.69       & 0.72$\pm0.005^\dag$        & & 0.83        & 0.76       & 0.79$\pm0.006^\dag$      &  &  0.63$\pm0.007$        & & 0.77$\pm0.004$              \\ 
Anatomy Subtype                       & 1,153                             & 0.70       & 0.64       & 0.67$\pm0.005^\dag$        & & 0.77        & 0.70       & 0.73$\pm0.006^\dag$      &  &  0.58$\pm0.006$        & & 0.71$\pm0.005$            \\ \bottomrule
\end{tabular}

%\begin{tabular}{m{1.4in} cccccccccccc}
%\toprule
%\multirow{3}{*}{\textbf{Span label}}  & \multirow{3}{*}{\textbf{\# gold}} & %\multicolumn{7}{c}{\textbf{SpERT}}                                                  &  & %\multicolumn{3}{c}{\textbf{BERT-SeqTag}}    \\ \cmidrule[\heavyrulewidth]{3-9} %\cmidrule[\heavyrulewidth]{11-13}
%                                      &                                   & \multicolumn{3}{c}{\textbf{exact}}       & & \multicolumn{3}{c}{\textbf{overlap}}   &  & \textbf{exact}       & & \textbf{overlap}   \\ \cmidrule{3-5} \cmidrule{7-9} \cmidrule{11-11} \cmidrule{13-13}
%                                      &                                   & \textbf{P} & \textbf{R} & \textbf{F1}    & & \textbf{P}  & \textbf{R} & \textbf{F1} &  & \textbf{F1}          & & \textbf{F1}        \\ \midrule
%Finding                               & 2,122                             & 0.82       & 0.83       & 0.83           & & 0.91        & 0.92       & 0.91        &  &  0.00                & & 0.00                    \\
%Anatomy                               & 1,153                             & 0.75       & 0.68       & 0.71           & & 0.83        & 0.76       & 0.79        &  &  0.00                & & 0.00                    \\ 
%Anatomy Subtype                       & 1,153                             & 0.70       & 0.64       & 0.67           & & 0.77        & 0.70       & 0.73        &  &  0.00                & & 0.00                    \\ \bottomrule
%\end{tabular}
        \caption{Span labeling performance}
        \label{span_performance}
    \end{subtable}

    \vspace{0.3in}

    \begin{subtable}[h]{6.5in}    
        \small
        \centering

\begin{tabular}{m{1.35in} cccccccccccc}
\toprule
\multirow{3}{*}{\textbf{Relations}}  & \multirow{3}{*}{\textbf{\# gold}} & \multicolumn{7}{c}{\textbf{SpERT}}                                                             &  & \multicolumn{3}{c}{\textbf{BERT-multi}}    \\ \cmidrule[\heavyrulewidth]{3-9} \cmidrule[\heavyrulewidth]{11-13}
                                      &                                   & \multicolumn{3}{c}{\textbf{exact}}          & & \multicolumn{3}{c}{\textbf{overlap}}          &  & \textbf{exact}       & & \textbf{overlap}   \\ \cmidrule{3-5} \cmidrule{7-9} \cmidrule{11-11} \cmidrule{13-13}
                                      &                                   & \textbf{P} & \textbf{R} & \textbf{F1}  & & \textbf{P}  & \textbf{R} & \textbf{F1}   &  & \textbf{F1}     & & \textbf{F1}        \\ \midrule
Finding-Anatomy                       & 1,380                             & 0.65       &  0.60      & 0.63$\pm0.005^\dag$      & & 0.75        & 0.70       & 0.72$\pm0.005^\dag$       & &  0.50$\pm0.005$         & & 0.66$\pm0.004$               \\
Finding-Anatomy Subtype               & 1,380                             & 0.61       &  0.56      & 0.58$\pm0.006^\dag$      & & 0.70        & 0.65       & 0.67$\pm0.005^\dag$       & &  0.47$\pm0.006$         & & 0.60$\pm0.006$             \\ \bottomrule
\end{tabular}

%\begin{tabular}{m{1.4in} cccccccccccc}
%\toprule
%\multirow{3}{*}{\textbf{Relations}}  & \multirow{3}{*}{\textbf{\# gold}} & \multicolumn{7}{c}{\textbf{SpERT}}                                                  &  & \multicolumn{3}{c}{\textbf{BERT-SentClass}}    \\ \cmidrule[\heavyrulewidth]{3-9} \cmidrule[\heavyrulewidth]{11-13}
%                                      &                                   & \multicolumn{3}{c}{\textbf{exact}}       & & \multicolumn{3}{c}{\textbf{overlap}}   &  & \textbf{exact}       & & \textbf{overlap}   \\ \cmidrule{3-5} \cmidrule{7-9} \cmidrule{11-11} \cmidrule{13-13}
%                                      &                                   & \textbf{P} & \textbf{R} & \textbf{F1}    & & \textbf{P}  & \textbf{R} & \textbf{F1} &  & \textbf{F1}          & & \textbf{F1}        \\ \midrule
%Finding-Anatomy                       & 1,380                             & 0.64       &  0.60      & 0.62           & & 0.75        & 0.69       & 0.72        & &  0.00                 & & 0.00               \\
%Finding-Anatomy Subtype               & 1,380                             & 0.60       &  0.56      & 0.58           & & 0.69        & 0.64       & 0.67        & &  0.00                 & & 0.00               \\ \bottomrule
%\end{tabular}

        \caption{Relation extraction performance}
        \label{relation_performance}
    \end{subtable}

    \vspace{0.1in}

    \caption{Average extraction performance on the withheld test set, as mean and standard deviation for 10 trained models. $^\dag$indicates best performance with significance ($p < 0.05$).}
    \label{extraction_performance}        
\end{table}

This section presents the entity and relation extraction performance. Tables \ref{span_performance} and \ref{relation_performance} presents the extraction performance on the withheld test set for SpERT and BERT-multi, averaged across 10 randomly instantiated models. Table \ref{span_performance} includes the span labeling performance for \textit{Finding} and \textit{Anatomy} entities and the micro-averaged \textit{Anatomy Subtype} labels. An \textit{Anatomy} label is assigned is any span with an \textit{Anatomy Subtype} label. SpERT outperforms BERT-multi for all span labels, under the \textit{exact match} or \textit{any overlap} criteria, with significance ($p<0.05$). There is a larger performance gap between the \textit{exact match} and \textit{any overlap} assessment for BERT-multi than SpERT, which is likely the result of the differing training objectives.  SpERT is trained to identify exact span matches without any reward for partial matches, while BERT-multi generates word piece predictions that are aggregated to token predictions. For both architectures, the relatively small difference in performance between \textit{Anatomy} and \textit{Anatomy Subtype} (0.04-0.05 $\Delta$F1 for \textit{exact match} and 0.06-0.07 $\Delta$F1 for \textit{any overlap}), suggests that there is relatively low confusability between the \textit{Anatomy Subtype} labels for spans that are correctly identified as \textit{Anatomy}.

%\end{wraptable}

\begin{wrapfigure}{R}{4.05in}
%\begin{figure}[ht]
    \small
    \centering
    \includegraphics[width=4.1in]{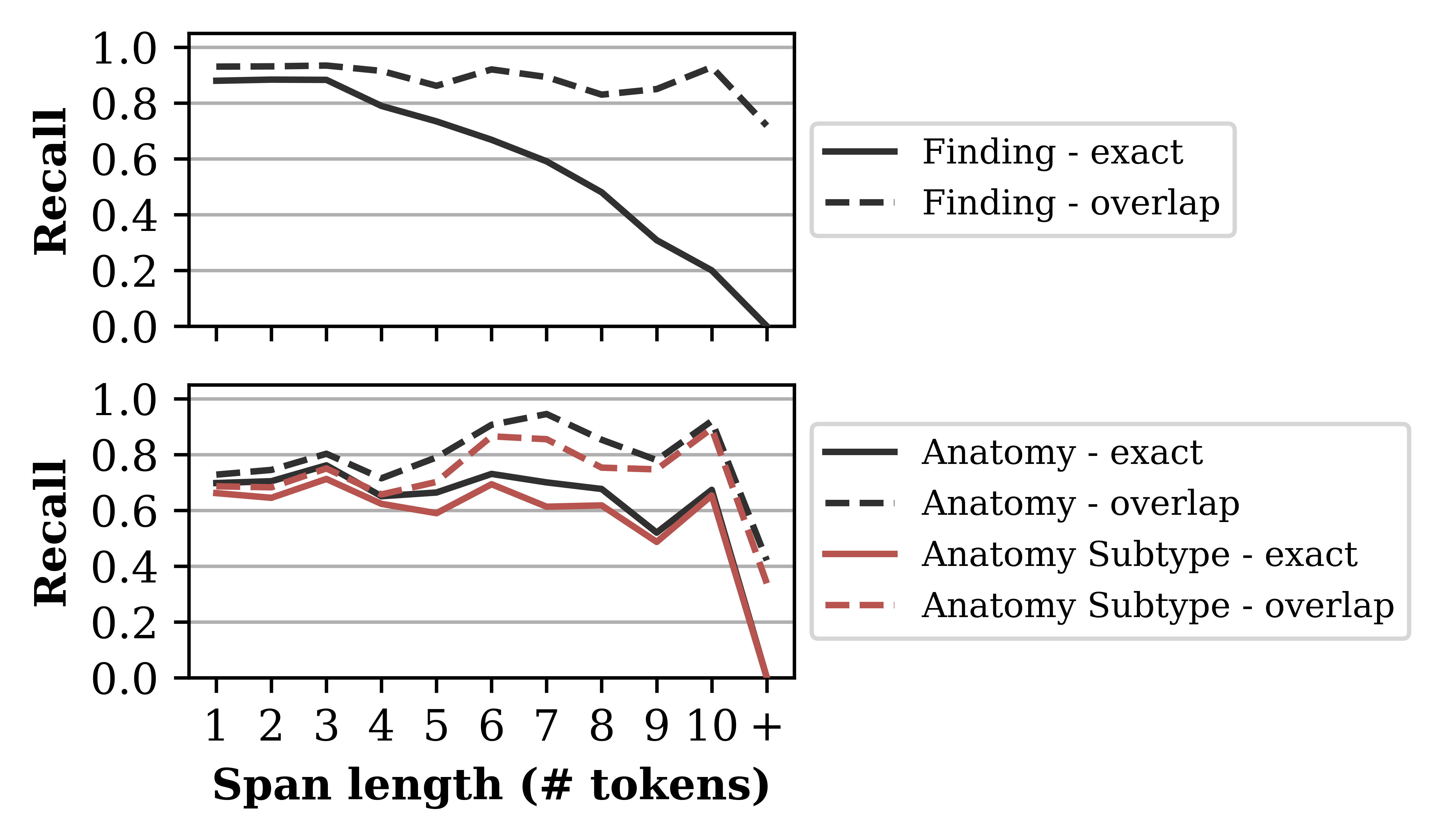}
    \caption{Span extraction recall as a function of span length in tokens (not word pieces)}
    \label{performance_vs_length}
%\end{figure}    
\end{wrapfigure}

Table \ref{relation_performance} presents the relation extraction performance. SpERT outperforms BERT-multi for \textit{Finding}-\textit{Anatomy} and \textit{Finding}-\textit{Anatomy Subtype} relations with significance. As expected, the relation extraction performance is lower than the span labeling performance because of cascading errors. For both architectures, the magnitude of the performance drop from span labeling to relation extraction is roughly consistent with the accumulation of \textit{Finding} and \textit{Anatomy} span labeling errors, suggesting that the performance of the relation classifiers is relatively high.

Figure \ref{performance_vs_length} presents the recall of SpERT as a function of the gold span length, in number of tokens (not word pieces). The recall is aggregated for the 10 model runs and reported for \textit{Finding}, \textit{Anatomy}, and and \textit{Anatomy Subtype} labels. The maximum span width for SpERT is set to 10 tokens, so the \textit{exact match} recall is zero for all spans longer than 10 tokens. Under the \textit{exact match} criteria, the \textit{Finding} recall drops from approximately 0.9 for shorter spans to approximately 0.2-0.3 for long spans (9-10 tokens). Under the \textit{any overlap} criteria, the \textit{Finding} recall remains relatively high for all span lengths, as the extractor only needs to identify a portion of the gold span for a match. Under the \textit{exact match} criteria, the \textit{Anatomy} and \textit{Anatomy Subtype} remains relatively steady across span lengths from 1-10. Under the \textit{any overlap} criteria, the \textit{Anatomy} and \textit{Anatomy Subtype} recall tends to increase with span length.

\begin{wrapfigure}{R}{4.0in}
%\begin{figure}[ht]
    \centering
    \includegraphics[width=4.0in]{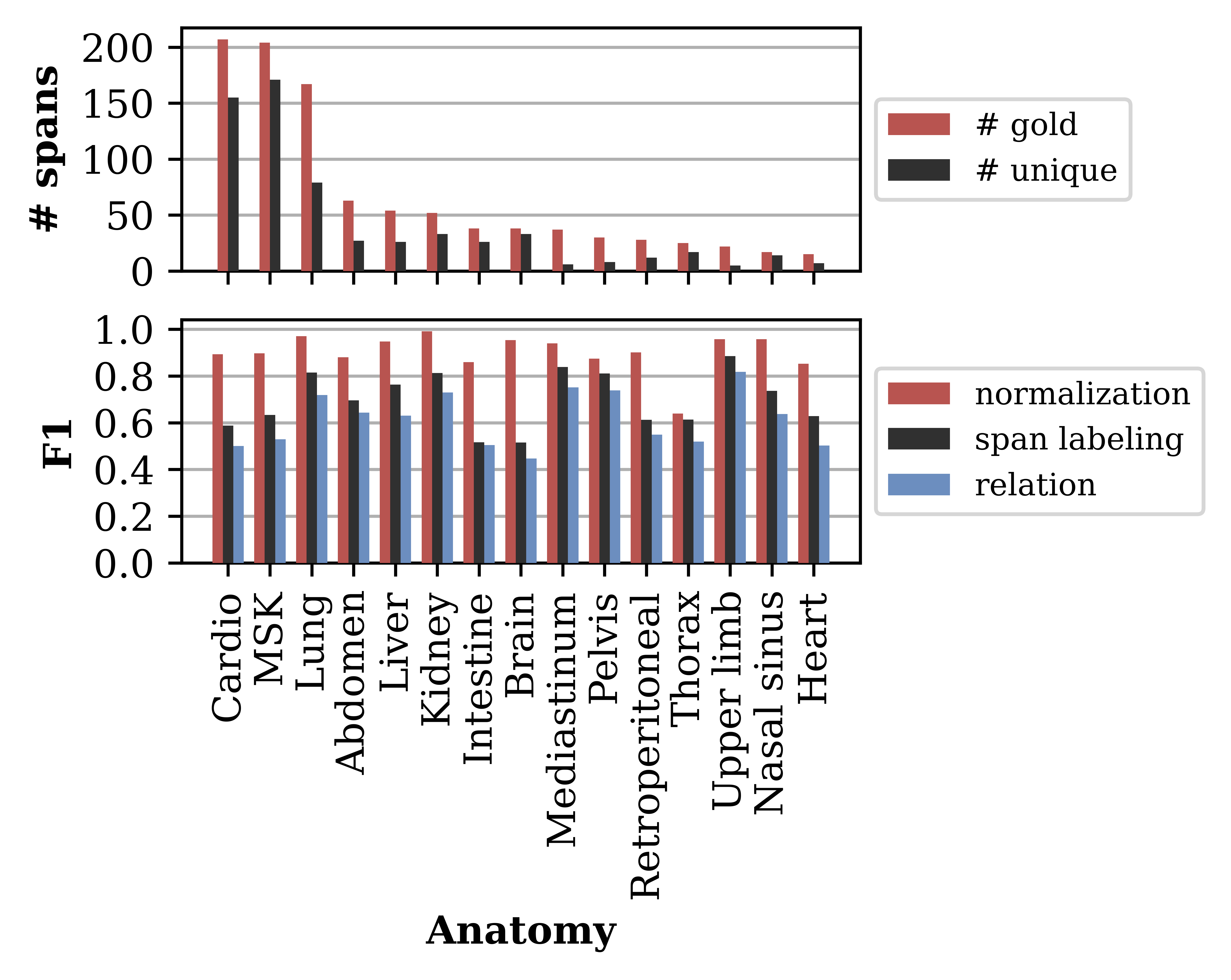}
    \caption{Performance, anatomy type frequency, and number of unique spans by \textit{Anatomy Subtype}.}
    \label{perf_by_anatomy}
%\end{figure}    
\end{wrapfigure}

Figure \ref{perf_by_anatomy} presents summary statistics and performance for the 15 most frequent \textit{Anatomy Subtypes} in the test set. Figure \ref{perf_by_anatomy} includes the label counts (\# gold), number of unique lower cased spans (\# unique). It also includes the normalization, span labeling, and relation extraction performance. The normalization performance is associated with \textit{sentence context} models summarized in Table \ref{normalization_performance}, and the span labeling and relation extraction performance is associated with the SpERT model summarized in Table \ref{extraction_performance}. There is a large imbalance in the distribution of \textit{Anatomy Subtype} labels with Cardio, MSK, and Lung accounting for approximately 50\% of the labels. The diversity of the anatomy spans varies significantly by \textit{Anatomy Subtype}. For example, MSK has 168 unique spans in 204 occurrences (ratio of 0.8), while Mediastinum has 7 unique spans in 37 occurrences (ratio of 0.2). The span labeling and relation extraction performance does not drop off for infrequent labels and appears to be more related to the span diversity.

\subsection*{Error Analysis}

\begin{wraptable}{r}{4.3in}
    %\vspace{1.0in}
%\begin{table}[ht]
    \small
    \centering
    \setlength\tabcolsep{2.5pt}

%\begin{tabular}{lllllllll}
\begin{tabular}{lll}
\toprule
\textbf{Span label}                  &\textbf{Short examples}   & \textbf{Long examples} \\ \midrule
\multirow{2}{*}{Finding}        & \say{hernia}             & \say{poor opacification of these vessels distally}          \\
                                & \say{lesion}             & \say{expanded thoracic aortic aneurysm}                     \\ \midrule
Anatomy                         & \say{scapula}            & \say{soft tissues of the posterolateral left chest wall}    \\ 
\hspace{0.1in}MSK               & \say{left third rib}     & \say{subcutaneous fat in the right groin}                   \\ \midrule
Anatomy                         & \say{aorta}              & \say{proximal descending thoracic aorta}                    \\
\hspace{0.1in}Cardio            & \say{coronary arteries}  & \say{arteries of the right lower extremity and abdomen}     \\ \midrule
Anatomy                         & \say{left lung}          & \say{lateral aspect of the right major fissure}             \\
\hspace{0.1in}Lung              & \say{right lower lobe}   & \say{dependent portions of the left upper lobe adjacent}    \\ \bottomrule
\end{tabular}
    \caption{Example false-negative spans}
    \label{false_negatives}
%\end{table}
\end{wraptable}

Generating a correct relation prediction requires identifying the \textit{Finding} (head), identifying the \textit{Anatomy} and \textit{Anatomy Subtype} (tail), and pairing the head and tail (role). The results in Table \ref{extraction_performance} suggests the biggest source of error is identifying \textit{Anatomy} entities, followed by identifying \textit{Finding} entities. Error is also introduced in the \textit{Anatomy Subtype} normalization and \textit{Finding}-\textit{Anatomy} pairing; however, entity extraction is the most challenging aspect of this task. Table \ref{false_negatives} presents example SpERT false negative spans for \textit{Finding} and the most frequent \textit{Anatomy Subtypes}. These false negatives are assessed using the \textit{any overlap} criteria, to identify text regions related to findings and anatomy that the model completely missed.

The short \textit{Finding} examples are relatively straightforward targets and the cause of these missed spans is unclear. The long \textit{Finding} examples include medical problems coupled with anatomical information, resulting in longer spans that are generally more difficult to extract. The inclusion of anatomical information in the \textit{Finding} spans creates annotation inconsistencies, where anatomical information may be labeled as \textit{Finding} or \textit{Anatomy}. We are currently building on this radiological work as part of an exploration of incidentalomas and updated the annotation guidelines to separate finding information from anatomy information and create shorter, more consistently annotated spans. For example the \textit{Finding} span, \say{expanded thoracic aortic aneurysm}, would be annotated as a the relation triple (\textit{Finding}=\say{aneurysm}, \textit{role}=\say{has}, \textit{Anatomy}=\say{thoracic aortic}).

All of the short \textit{Anatomy} examples are concise descriptions of anatomy that use common anatomical terminology. There are multiple contributing factors to these errors. In the corpus, only anatomy information associated with findings is annotated, so there are many descriptions of anatomy that are not annotated. As previously discussed, anatomy information is frequently incorporated into \textit{Finding} annotations, which introduces annotation inconsistencies. The long \textit{Anatomy} examples are more nuanced descriptions of anatomy that often describe multiple systems or body parts in relation to each other. For example, the Cardio span, \say{arteries of the right lower extremity and abdomen}, contains references of three \textit{Anatomy Subtypes}: Cardio, Lower limb, and Abdomen. Annotating such examples with the \textit{Anatomy Subtype} labels can be challenging, and more nuanced anatomy descriptions are likely to have noisier annotations.

\ifsubfile
\bibliography{mybib}
\fi

\section*{Conclusions}

This work explores a novel radiological information extraction task with the goal of automatically generating semantic representations of radiological findings that capture anatomical information. We extract and normalize anatomical information connected to findings in CT reports, using state-of-the-art extraction architectures. This extraction task is both novel and important because it couples extracted anatomical information with radiological findings and normalizes the anatomical information to a commonly used ontology. Linking the anatomy to findings and normalizing the anatomy yields a more complete semantic representation, which can more easily be incorporated into secondary use applications. We demonstrate that the span-based SpERT model, which jointly extracts entities and relations, outperforms a strong BERT baseline that separately extracts entities and relations in a pipelined approach. The explored extraction task involves three subtasks: identifying \textit{Finding} and \textit{Anatomy} entities, normalizing \textit{Anatomy} entities to \textit{Anatomy Subtypes}, and pairing \textit{Finding} and \textit{Anatomy} entities through relations. Entity extraction is the most difficult of these subtasks. We find that extraction performance for \textit{Finding} entities decreases as span length increases; however, \textit{Anatomy} extraction performance is relatively constant across span lengths. In an exploration of performance by \textit{Anatomy Subtype}, we find span extraction performance is influenced more by the diversity of the associated spans than the frequency of the \textit{Anatomy Subtype} labels.

This work is limited by the annotated data set, which only utilizes data from a single hospital system and incorporates a single type of imaging report (CT). The extraction models trained on this annotated data set may not generalize well to other institutions or radiology modalities. We are currently expanding the annotated data set to other radiology modalities, including magnetic resonance imaging (MRI) and positron emission tomography (PET) reports. 

The 56 \textit{Anatomy Subtypes} used in this work provide moderate granularity in resolving the anatomical location of radiological findings. In our current incidentaloma research, we anticipate representing anatomical locations with finer resolution. We will build on the work presented here and explore learned approaches for characterizing anatomical spans through multiple attributes. For example the phrase \say{right lower lobe} could be characterized through a semantic representation describing the body part/organ (Lung), sidedness (right), and vertical location (lower). This type of detailed semantic representation could facilitate a wide range of impactful use cases.

\ifsubfile
\bibliography{mybib}
\fi

%\clearpage
\section*{Acknowledgements}
This work was supported by NIH/NCI (1R01CA248422-01A1) and NIH/NLM (Biomedical and Health Informatics Training Program - T15LM007442). Research and results reported in this publication were partially facilitated by the generous contribution of computational resources from the University of Washington Department of Radiology.

\bibliography{mybib}

\end{document}